\documentclass[10pt,twocolumn,letterpaper]{article}

\usepackage{wacv}
\usepackage{times}
\usepackage{epsfig}
\usepackage{graphicx}
\usepackage{amsmath}
\usepackage{amssymb}
\usepackage{placeins}

\def\wacvPaperID{410} 

\wacvfinalcopy 

\ifwacvfinal
\def\assignedStartPage{1} 
\fi

\ifwacvfinal
\usepackage{hyperref}
\hypersetup{breaklinks=true,bookmarks=false}
\else
\usepackage{hyperref}
\hypersetup{pagebackref=true,breaklinks=true,colorlinks,bookmarks=false}
\fi

\ifwacvfinal
\else
\fi

\begin{document}

\title{Making DensePose fast and light}

\author{
    Ruslan Rakhimov\textsuperscript{1}\thanks{Equal contribution}\space, 
    Emil Bogomolov\textsuperscript{1}$^*$, 
    Alexandr Notchenko\textsuperscript{1},\\
    Fung Mao\textsuperscript{2}, 
    Alexey Artemov\textsuperscript{1}, 
    Denis Zorin\textsuperscript{1,3}, 
    Evgeny Burnaev\textsuperscript{1}\\
    \textsuperscript{1}{\small Skolkovo Institute of Science and Technology}\\
    \textsuperscript{2}{\small Huawei Moscow Research Center (Russia)}\\
    \textsuperscript{3}{\small New York University}\\
    {\tt\small\{ruslan.rakhimov, e.bogomolov, alexandr.notchenko\}@skoltech.ru,}\\
    {\tt\small fung.mao@huawei.com, a.artemov@skoltech.ru, dzorin@cs.nyu.edu, e.burnaev@skoltech.ru}\\
}

\maketitle


\begin{abstract}
DensePose estimation task is a significant step forward for enhancing user experience computer vision applications ranging from augmented reality to cloth fitting. Existing neural network models capable of solving this task are heavily parameterized and a long way from being transferred to an embedded or mobile device. To enable Dense Pose inference on the end device with current models, one needs to support an expensive server-side infrastructure and have a stable internet connection. To make things worse, mobile and embedded devices do not always have a powerful GPU inside.
In this work, we target the problem of redesigning the DensePose R-CNN model's architecture so that the final network retains most of its accuracy but becomes more light-weight and fast. To achieve that, we tested and incorporated many deep learning innovations from recent years, specifically performing an ablation study on 23 efficient backbone architectures, multiple two-stage detection pipeline modifications, and custom model quantization methods. 
As a result, we achieved $17\times$ model size reduction and $2\times$ latency improvement compared to the baseline model.\ifwacvfinal\footnote{Code is available at \url{https://github.com/zetyquickly/DensePoseFnL}}\fi
\end{abstract}
\section{Introduction}

This work is dedicated to developing an architecture for solving DensePose \cite{densepose} estimation task with a particular requirement: the model should be light-weight and run in real-time on a mobile device.

The task of understanding humans in an image may involve different formulations of the problem: 2d landmarks localization, human part segmentation, 3d reconstruction, dense image-to-surface correspondences (DensePose). In this work, we target the multi-person formulation of DensePose task: given a single RGB image solve the regression task: for each pixel, find its surface points (UV coordinates) on a deformable surface model (the Skinned Multi-Person Linear (SMPL) model \cite{smpl}).

Finding surface correspondence is a step forward to a general 3d human representation. Possible applications lie in such fields, like augmented reality, virtual fitting rooms. Densepose output may serve as input to another model. For instance, it was used as an input in video-to-video translation tasks \cite{vid2vid}.

Besides the original pioneering work \cite{densepose}, which introduces a carefully annotated COCO-DensePose dataset with sparse image-to-surface ground-truth correspondences and DensePose R-CNN baseline model, other works target different formulations. Parsing R-CNN \cite{parsing}, the winner solution of the COCO 2018 Challenge DensePose Estimation task, achieves state-of-the-art performance by scrutinizing different blocks in the original DensePose R-CNN architecture. Slim DensePose \cite{denseposeslim} explores the weakly-supervised and self-supervised learning problem setting, by leveraging motion cues from videos. \cite{uncertainty} improves the performance of the model by incorporating the uncertainty estimation into the model. \cite{monkeys} shows the ability to transfer the dense pose recognition from humans to proximal animal classes such as chimpanzees without a time-consuming collection of a new dataset with new classes.

However, none of the works target the task of making the network fast and light-weight, and current solutions such as baseline DensePose R-CNN and state-of-the-art Parsing R-CNN introduce heavily parametrized models.

Make the network perform near to a real-time mode is a particularly important step if we want to apply these models in the mobile or embedded devices. In this work, we explore the subtle trade-off between the performance of the model and its latency.

The contributions are the following:
\begin{itemize}
    \item we created a pipeline to test neural network architectures viability for mobile deployment,
    \item we developed an architecture based on existing techniques, achieving a finally good balance between real-time speed and average precision of our model,
    \item we performed an ablation study on many different efficient backbones, particularly applied for DensePose task.
\end{itemize} 
\section{Related work}

\noindent 
\textbf{DensePose task.} DensePose-COCO dataset contains a large set of images of people collected ``in the wild'' together with different annotations: (i) bounding boxes, (ii) foreground-background masks, (iii) dense correspondences --- points $p \in S$ of a reference 3D model $S\in\mathbb{R}^3$ of the object associated with triplets $(c, u, v) \in\{1, \ldots, C\} \times[0,1]^{2}$, where $c$ indicates which one of $C$ body parts contains the pixel and $(u,v)$ represents the corresponding location (UV coordinates) in the chart of the part \cite{smpl}.
The DensePose task is then to predict such triplets $(c, u, v)$ for each foreground pixel and every person in the image.
\newline

\noindent \textbf{DensePose R-CNN.} The baseline dense pose prediction model, and all the subsequent works \cite{parsing, uncertainty, monkeys} follow the architecture design of Mask R-CNN \cite{maskrcnn}.

\begin{figure}[!hbtp]
\centering
\includegraphics[width=0.4\textwidth]{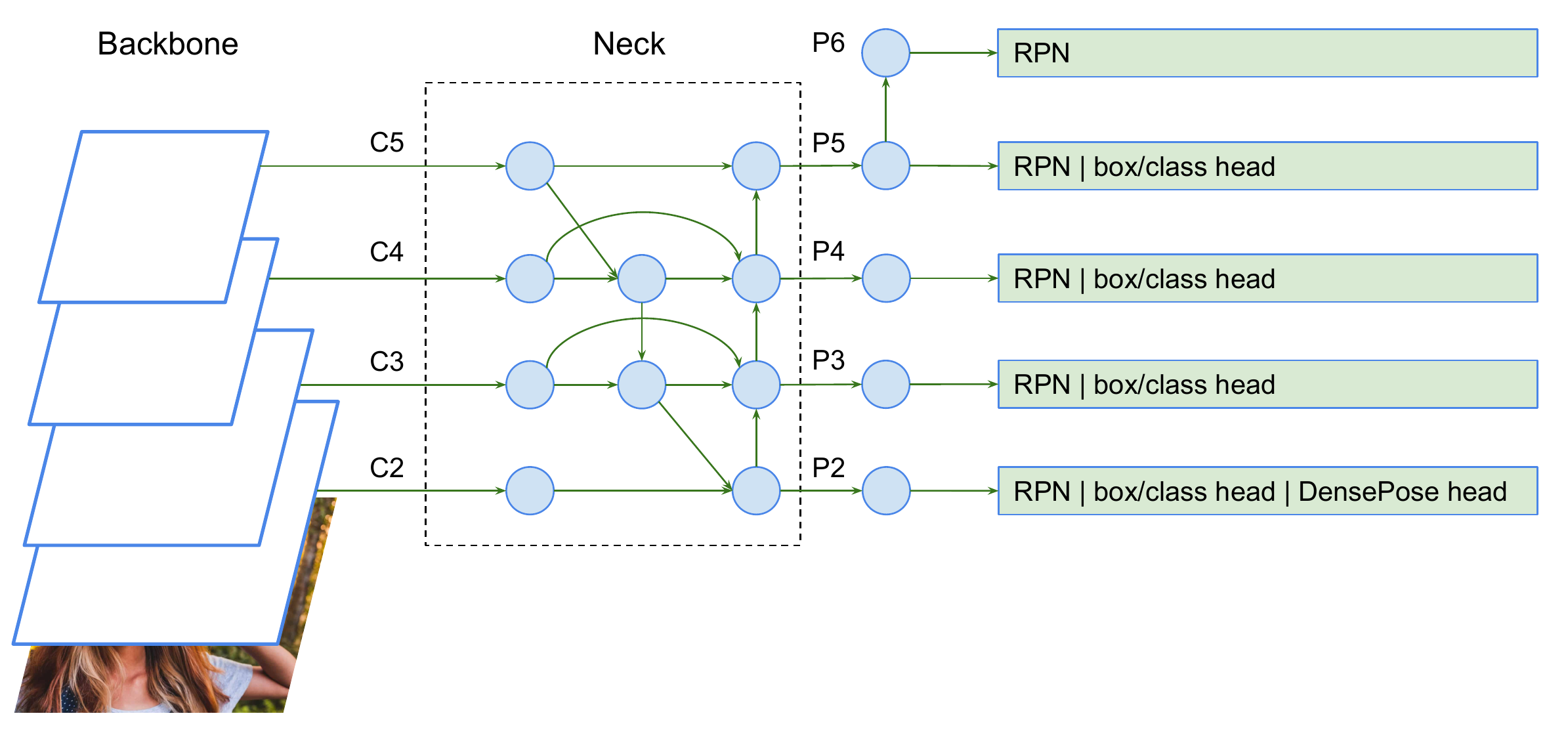}
\caption{The high level structure of the Mobile Parsing R-CNN model. $C_i$, $P_i$ represent feature levels with a resolution of $1/2^i$ of the input image. $P_6$ is obtained via stride-2 pooling on $P_5$.}
\label{fig:scheme}
\end{figure}

The model is a two-stage: first, it generates class-independent region proposals (boxes), then classifies and refines them using the box/class head. Finally, the DensePose head predicts the body part and UV coordinates for each pixel inside the box. Particularly, the model consists of many different blocks (see Fig.~\ref{fig:scheme}):
\begin{itemize}
    \item \textit{Backbone} to extract features from the image,
    \item \textit{Neck} to integrate features from different feature levels of the backbone to effectively perform multi-scale detection,
    \item \textit{Region proposal network (RPN)} to propose a sparse set of box candidates potentially containing objects,
    \item \textit{Heads} take the features pooled from the bounding box on the corresponding feature level, where the detection occurred, and produce output. The first head is a box/class head, which finally predicts whether the object is present in the box and refines the box coordinates. The second head is the DensePose head that predicts either the pixel belongs to the background or assigns it to one of the 24 DensePose charts, and regresses UV coordinates to each foreground pixel inside the bounding box.
\end{itemize}

\noindent \textbf{Model architecture optimisation.}
In recent years the neural architecture search (NAS) techniques gained popularity \cite{automl}. The main aim of NAS is to find the optimal architecture under specific hardware requirements. Usually, these techniques are applied in simple setups, e.g., classification networks, or in the case of two-stage object detection models, NAS is usually applied to individual parts of the model \cite{nasfpn}. In this paper, instead of creating one more design for a particular part of the model, we try to test different existing approaches and see what works best for the DensePose estimation task. Particularly, we evaluate several backbones that were a result of NAS optimization and try to test them out with other components.

\section{Mobile Parsing R-CNN}
\label{section3}

In this section, we address the design choice of different parts inside our model, which we call Mobile Parsing R-CNN. In general, the model's design follows the Parsing R-CNN model, the winner solution of the COCO 2018 Challenge DensePose Estimation task, but with different modifications in different parts.

\subsection{Backbone}

While there are many different possible designs of a backbone network, we target efficient models with a block structure as that in MobileNetV1 and V2 \cite{mobilenetv1, mobilenetv2} (depth-wise separable convolutions and inverted residuals with linear bottlenecks). This base block is the foundation for most efficient backbones used today, which were selected for evaluation as the backbone of the improved model. 
Let us list various architectures we use in our experiments:
\begin{itemize}
    \item \textbf{MobileNetV3}.  \cite{mobilenetv3} applies neural architecture search (NAS) and improves MobileNetV2 by adopting Squeeze and excitation block for channel-wise attention and non-linearities like h-sigmoid and h-swish;
    \item \textbf{MixNet}. \cite{mixnet} develops a multi-kernel variant of MobileNetV2, i.e., depth-wise convolutions consisting of convolutions with different kernel sizes;
    \item \textbf{Differentiable NAS} considers the problem of finding neural architecture in a differentiable way by carefully designing search space. We consider the following models, obtained using the differentiable NAS procedure: MnasNet \cite{mnasnet}, FBNet \cite{fbnet}, Single-Path\cite{spnasnet};
    \item \textbf{EfficientNets} from \cite{efficientnetb} appear to be one of the first architectures, obtained using AutoML approaches for image classification, and achieve a good compromise between the accuracy on a classification task and the number of the parameters of the network. \cite{efficientnetb} shows that one can apply a power-law scaling of width as a function of depth. Later EfficientNets were customized for Google's Edge TPUs \cite{efficientnete} using MNAS framework \cite{mnasnet};
    \item \textbf{CondConv}. Traditional convolutional layers have the kernel weights fixed once they are trained. CondConv \cite{condconv} applies a linear combination of several kernels (a mixture of experts) with weights generated dynamically by another network based on the input. While the original work is devoted to the classification task, we explore this \textit{``dynamic''} approach combined with EfficientNets on the DensePose task.
\end{itemize}

\subsection{Neck}

The main challenge in the object detection pipeline is to be able to detect objects of different scales. Earlier detectors predict objects based on features extracted from different levels of the backbone. Later, feature pyramid network (FPN \cite{fpn}) proposes to integrate features in a top-down manner to enrich fine-grained features from the lowest level of feature pyramid with semantically rich information from deeper layers. While the original work \cite{fpn} considers only the top-down pathway for information aggregation, later works also add cross-scale connections between the feature levels. In this work, we make use of bidirectional FPN (BiFPN \cite{bifpn}) for multi-scale feature fusion, which outperforms its recent counterparts in object detection tasks (see \cite{bifpn}), while remaining light-weight and fast. It is partly achieved by using separable convolutions inside.

\subsection{Densepose head}

We increase the region of interest (RoI) resolution for the DensePose head from $14\times14$ to $32\times32$, as it was suggested in \cite{parsing}.

While the original network uses 8 convolutions layers in the DensePose head, we, instead, similar to \cite{parsing}, use the atrous spatial pyramid pooling (ASPP) \cite{aspp} module, followed by 4 convolutional layers. Also, we omit the non-local convolutional layer \cite{nonlocal} between ASPP and convolutional layers in order not to increase the latency of the network because it performs pixel to pixel comparisons resulting in $O(n^2)$ operations, where $n$ is the number of pixels.

Finally, the DensePose predictions happen on the finest level from the feature pyramid as in \cite{parsing}, while box/class predictions happen on all levels.

\subsection{Quantization of backbone layers}
\label{quant}

We proposed the quantization procedure for Parsing R-CNN based on quantization aware training tools provided by PyTorch. 
First of all, it is necessary to patch the existing network architecture. Considering the whole network operates with quantized tensors, we should find intermediate parts where floating-point tensors are crucial to obtain satisfactory results.
\begin{enumerate}
    \item RPN classification and regression heads use a $3\times3$ convolutional layer to produce a shared hidden state from which one $1\times1$ convolutional layer predicts objectness logits for each anchor, and another one predicts bounding-box deltas specifying how to refine the anchor coordinates to get a final object proposal. These layers work with quantized feature tensors, but for correct calculation of RPN proposals, predicted objectness logits and anchor deltas are dequantized after inference of bounding box predictor.
    \item To perform accurate RoI pooling, it is necessary first to dequantize input features, apply pooling, and then quantize features back.
\end{enumerate}

The second step is fusion. We fuse each convolutional and linear layer, followed by batch normalization and activation to one atomic layer. That is needed to save on memory access while also improving the operations' numerical accuracy.
The third step is to run the quantization aware training of the patched and fused model.

During the second and third steps, we run into design obstacles that are described below.

\begin{figure}[!hbtp]
\centering
\includegraphics[width=0.4\textwidth]{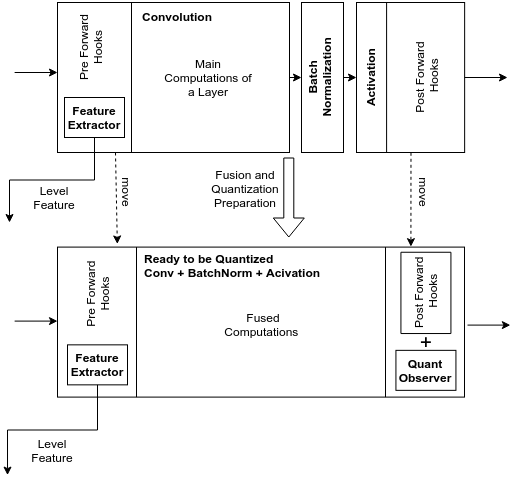}
\caption{The feature collection scheme for quantized models.}
\label{fig:preserve_hooks}
\end{figure}

In BiFPN architecture, we collect features before point-wise linear convolutions using pre-forward hooks. This allows us to link to this layer's input rather than to the output of the input provider. But quantization tools implemented in the PyTorch framework at this stage do not allow this to be done. We proposed a mechanism that preserves pre- and post- forward hooks during fusion and preparation for quantization and does not harm the quality of the quantization process itself. The diagram of the proposed mechanism is in Fig.~\ref{fig:preserve_hooks}.

\section{Experiments}

In this section, we provide the experimental results on design choices for different parts of the model. The majority of experiments are done on the cluster, and finally, we transfer the model to a mobile device to check the performance there.

\subsection{Implementation details}

The models were implemented in PyTorch using Detectron2 \cite{detectron2} platform.

We choose hyper-parameters matching to those in Parsing R-CNN\cite{parsing}, i.e., we use a batch of 16 images (2 images per GPU), therefore we apply synchronous batch normalization \cite{syncbn} instead of usual batch normalization wherever it is used inside the backbone and neck. We use no normalization in box/class and dense pose heads. We sample 512 RoIs for box/class head and 32 RoIs for dense pose head. By default, we train models for 130k iterations with initial learning rate 0.002, decreasing it by 10 at 100k and 120k iterations. Under such a schedule, training of one model takes approximately 1 day on 8 NVIDIA Tesla V100 GPUs. Since all models are quite small, the memory consumption during training allows to decrease the number of GPUs for parallel training. Unless specified otherwise, by default, we scale images in a way that shortest image size equals 800 pixels during the inference stage. Each model's backbone is initialized with weights of the corresponding network trained on the ImageNet classification task. We train models on a combination of \textit{train} and \textit{valminusminival} partitions of Densepose-COCO dataset \cite{densepose} and test them on a \textit{minival} partition.

\subsection{Metrics}
Following the original work, we use as evaluation metric the Average Precision (AP) at a number of \textit{geodesic point similarity} (GPS) thresholds ranging from 0.5 to 0.95. We also report box average precision.

As we are interested in deploying a DensePose model on a mobile device, we report the number of parameters of each model and FPS measured on CPU and GPU. In particular, we measure the inference performance of all models on the same NVIDIA GeForce GTX 1080 Ti. It is worth mentioning that the DensePose model is a two-stage model, so the FPS of the model is directly conditioned on the performance of the first stage of the model. There is a subtle trade-off between the quality and latency; as for example, the network that does not predict any instances will never run dense pose head, and vice versa, the network that produces many false-positive results would redundantly run the model heads. As the latency of the network is data-dependent, we average latency time across DensePose-COCO minival dataset and finally convert it to FPS.

\subsection{Ablation on components}

\begin{table*}[t!]
\centering
\resizebox{\textwidth}{!}{
\begin{tabular}{cllll}
 &  DensePose R-CNN (baseline) \cite{densepose} & Parsing R-CNN \cite{parsing} & Mobile Parsing R-CNN (A) & Mobile Parsing R-CNN (B) \\\hline
Backbone & ResNet-50 \cite{resnet} & ResNet-50 \cite{resnet} & Single-Path \cite{spnasnet} & Single-Path \cite{spnasnet} \\
Neck & FPN\cite{fpn} & FPN\cite{fpn} & FPN\cite{fpn} & BiFPN  \\
RoI resolution & $14\times14$ & $32\times32$ & $32\times32$ & $32\times32$ \\
Pooling Type & RoIPool & RoIPool & RoIAlign & RoIAlign \\
Box/class head & 2 linear layers & 2 linear layers & 2 conv layers & 2 conv layers \\
Feature level for prediction & $P_2$,$P_3$,$P_4$,$P_5$ & $P_2$ & $P_2$ & $P_2$ \\
DensePose head & 8 conv layers & ASPP\cite{aspp}+NL\cite{nonlocal}+4 conv layers & ASPP\cite{aspp}+4 conv layers & ASPP\cite{aspp}+4 conv layers \\
\#Channels & 512 & 512 & 256 & 64 \\
\hline
\#Params & 59.73M & 54.36M & 11.35M & 3.35M \\
GPU FPS & $13.16$ & $10.15$ & $12.03$ & $22.77$ (3x LR: $\mathbf{23.55}$) \\
CPU FPS & $1.62$ & $1.39$ & $1.42$ & $2.02$  (3x LR: $\mathbf{2.10}$) \\
box AP & $57.8$ & $59.609$ & $56.370$ & $55.39$ (3x LR: $56.83$)\\
densepose AP & $49.8$ & $54.676$ & $49.512$ & $46.79$ (3x LR: $51.08$)\\
\hline
\end{tabular}}
\caption{The main differences between the models presented. Results on DensePose-COCO minival. 3x LR refers to 3 times longer training compared to the default setting. $P_i$ represents a feature level with a resolution of $1/2^i$ of the input images. \#Channels represent the number channels inside \textit{neck} and \textit{heads}.}
\label{table:configs}
\end{table*}



\begin{table*}[t!]
\centering
\resizebox{\textwidth}{!}{
\begin{tabular}{lcccccc}
\hline
Backbone & Top-1 Accuracy (\%) & \#Params & box AP & dp. AP & GPU FPS & CPU FPS \\ \hline
ResNet-50 \cite{resnet} & $77.15$ & $33.61$M & $60.0$ & $\mathbf{54.7}$ & $11.05$ & $1.34$ \\
EfficientNet-B3 \cite{efficientnetb} & $81.636$ & $16.03$M & $59.027$ & $53.084$ & $8.31$ & $1.37$ \\
EfficientNet-EdgeTPU-L \cite{efficientnete} & $80.534$ & $17.89$M & $60.069$ & $53.378$ & $8.11$  & $1.34$ \\
MixNet-XL \cite{mixnet} & $80.120$ & $19.10$M & $58.444$ & $51.475$ & $8.54$  & $1.32$ \\
EfficientNet-B2 \cite{efficientnetb} & $79.688$ & $13.68$M & $58.041$ & $51.800$ & $9.33$ & $1.38$ \\
MixNet-L \cite{mixnet} & $78.976$ & $14.62$M & $57.481$ & $50.649$ & $8.52$ & $1.34$ \\
EfficientNet-EdgeTPU-M \cite{efficientnete} & $78.742$ & $14.57$M & $58.825$ & $52.302$ & $9.21$ & $1.37$ \\
EfficientNet-B1 \cite{efficientnetb} & $78.692$ & $13.03$M & $57.654$ & $51.053$ & $9.49$ & $1.39$ \\
CondConv-EfficientNet-B0 \cite{efficientnete, condconv} & $77.304$ & $18.32$M & $56.779$ & $49.231$ & $10.63$ & $1.40$ \\
EfficientNet-EdgeTPU-S \cite{efficientnete} & $77.264$ & $13.12$M & $58.296$ & $51.606$ & $10.03$ & $1.39$ \\
MixNet-M \cite{mixnet} & $77.256$ & $12.39$M & $56.834$ & $48.371$ & $9.39$ & $1.35$ \\
EfficientNet-B0 \cite{efficientnetb} & $76.912$ & $12.10$M & $56.271$ & $49.647$ & $10.53$ & $1.39$ \\
MixNet-S \cite{mixnet} & $75.988$ & $11.52$M & $55.132$ & $46.685$ & $10.34$ & $1.37$ \\
MobileNetV3-Large-1.0 \cite{mobilenetv3} & $75.516$ & $12.04$M & $54.537$ & $47.195$ & $11.54$ & $1.40$ \\
MnasNet-A1 \cite{mixnet} & $75.448$ & $10.94$M & $54.648$ & $47.036$ & $11.21$ & $1.38$ \\
FBNet-C \cite{fbnet} & $75.124$ & $11.49$M & $55.399$ & $47.983$ & $10.97$ & $1.37$ \\
MnasNet-B1 \cite{mnasnet} & $74.658$ & $11.31$M & $52.280$ & $47.658$ & $11.24$ & $1.37$ \\
Single-Path \cite{spnasnet} & $74.084$ & $11.35$M & $56.370$ & $49.512$ & $\mathbf{12.03}$ & $\mathbf{1.42}$ \\
MobileNetV3-Large-0.75 \cite{mobilenetv3} & $73.442$ & $10.92$M & $52.763$ & $44.736$ & $11.02$ & $1.36$ \\
MobileNetV3-Large-1.0 (minimal) \cite{mobilenetv3} & $72.244$ & $10.48$M & $52.464$ & $44.632$ & $11.33$ & $1.36$ \\
MobileNetV3-Small-1.0 \cite{mobilenetv3} & $67.918$ & $10.07$M & $49.614$ & $35.808$ & $10.62$ & $1.35$ \\
MobileNetV3-Small-0.75 \cite{mobilenetv3} & $65.718$ & $9.74$M & $44.224$ & $32.650$ & $10.16$ & $1.33$ \\
MobileNetV3-Small-1.0 (minimal) \cite{mobilenetv3} & $62.898$ & $9.58$M & $45.989$ & $36.522$ & $10.34$ & $1.34$ \\\hline
\end{tabular}}
\caption{Ablation on the backbone network used in Mobile Parsing R-CNN (A). The backbones are sorted by top-1 accuracy. Results on DensePose-COCO \textit{minival}}
\label{table:backbones}
\end{table*}

\begin{figure*}[t!]
\centering
\includegraphics[width=0.9\textwidth]{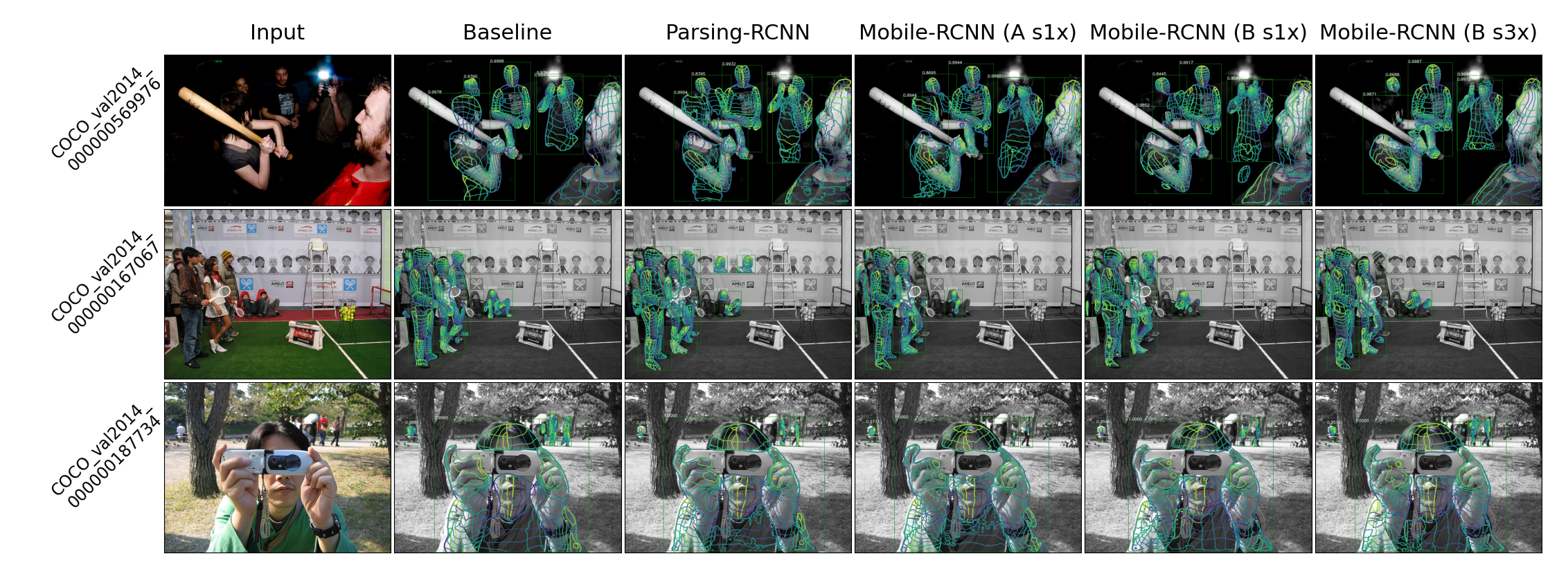}
\caption{Qualitative comparison of different models. We depict contours with color-coded U and V coordinates as an output of the model.}
\label{fig:fconfigs}
\end{figure*}

First, we implemented the Parsing R-CNN \cite{parsing} in Detectron2 following the original implementation. Then we modify the architecture exploiting the techniques presented in the Section~\ref{section3} and present two versions of a new model: Mobile Parsing R-CNN (A) and Mobile Parsing R-CNN (B). See the main architecture differences and obtained results in Table~\ref{table:configs}. Parsing R-CNN outperforms the baseline DensePose R-CNN model by $4.9$ AP, while the Mobile Parsing R-CNN (A) becomes more light-weight with the densepose AP similar to that achieved by the baseline model. The qualitative comparison can be seen in Fig.~\ref{fig:fconfigs}.

Specifically, Mobile Parsing R-CNN (A) is evolved from Parsing R-CNN by careful choice of a backbone, removing non-local block \cite{nonlocal}, decreasing the number of channels in FPN and all heads. Finally, we replace linear layers with convolutional ones in a box/class head. The results of a backbone comparison for Mobile Parsing R-CNN (A) can be seen in Table~\ref{table:backbones}. We use the backbones pretrained on ImageNet from \cite{geffnet}. First, we see that ResNet-50 provides a solid baseline both in terms of AP and FPS. The good FPS can be explained by the fact that the ResNet-50 is one of the first widespread popular deep networks, and GPU manufacturers constantly include this model for bench-marking. In the meantime, other networks contain specific new custom layers and are mainly designed for mobile or embedded devices. Nevertheless, by analyzing results in the Table~\ref{table:backbones}, we pick the Single-Path \cite{spnasnet} backbone as a network providing a good balance between FPS and the dense pose AP.

We move on from Mobile Parsing R-CNN (A) to Mobile Parsing R-CNN (B), by introducing a  new feature aggregation module (BiFPN \cite{bifpn} instead of FPN \cite{fpn}) and further decreasing the number of channels in the dense pose head by a factor of 4, thus 8 times lower than in the baseline architecture. The individual effects of each change can be found in Table~\ref{table:channels}. The transfer from FPN to BiFPN results in a reduced number of parameters,  better box, and densepose AP and the identical FPS. The $4\times$ decrease of the number of channels in BiFPN and all heads results only in $6.0$ densepose AP reduction, while increasing FPS approximately $2$ times.

Our implementation of BiFPN differs from the original one in terms of up-sampling and down-sampling procedure type used to make features from different levels of backbone spatially compatible for the fusion. While the original work uses a bilinear (up)down-sampling, we use a nearest-neighbor variant since we found it to be much faster on mobile devices, and the drop in AP is very slight. Also, in the case of BiFPN, we use features before point-wise linear $1\times1$ convolutions, compared to ``after'' in case of FPN, as it results in slight improvement of dense pose AP. Finally, we train the model three times more iterations, i.e., 390k iterations, reducing learning rate by 10 at 330k and 370k iterations and call it Mobile Parsing R-CNN (B s3x).

\begin{table*}[t!]
\centering
\resizebox{0.8\textwidth}{!}{
\begin{tabular}{cccccccc}
\hline
 & Neck & \#channels & \#Params & box AP & dp. AP & GPU FPS & CPU FPS  \\\hline
Mobile Parsing R-CNN (A) & FPN & $256$ & $11.35$M & $56.371$ & $49.512$ & $12.03$ & $1.42$ \\
 & BiFPN & $256$ & $10.53$M & $58.106$ & $52.80$ & $12.05$ & $1.41$ \\
& BiFPN & $112$ & $4.41$M & $56.41$ & $49.64$ & $19.04$ & $1.78$ \\
& BiFPN & $88$ & $3.82$M & $56.08$ & $48.19$ & $20.43$ & $1.87$ \\
Mobile Parsing R-CNN (B) & BiFPN & $64$ & $3.35$M & $55.39$ & $46.79$ & $22.77$ & $2.02$ \\\hline
\end{tabular}}
\caption{Ablation on neck type and number of channels. The number of channels is the same in neck and heads. Results on DensePose-COCO \textit{minival}}
\label{table:channels}
\end{table*}

\begin{figure*}[t!]
\centering
\includegraphics[width=0.8\textwidth]{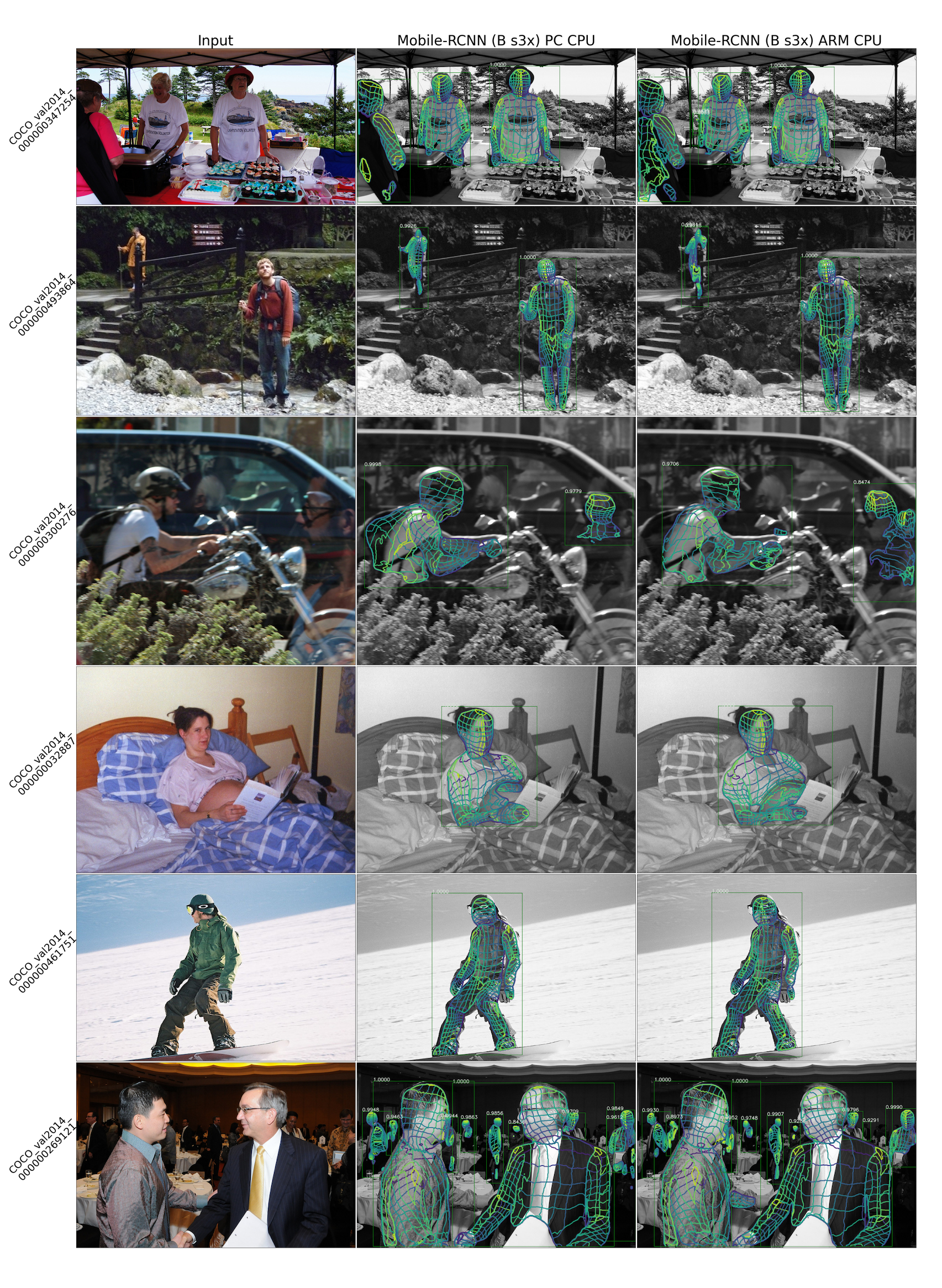}
\caption{Qualitative comparison of different backends. We depict contours with color-coded U and V coordinates as an output of the model.}
\label{fig:configs}
\end{figure*}

\subsection{Smartphone-based implementation}

\begin{table*}[t!]
\centering
\resizebox{0.8\textwidth}{!}{
\begin{tabular}{cccccc}
\hline
image shortest side & box AP & dp. AP & GPU FPS & CPU FPS & mobile CPU FPS\\\hline
200px & 36.449 & 19.028 & 27.549 & 10.277 & 2.355\\
400px & 49.181 & 43.916 & 24.648 & 6.921 & 0.954\\
512px & 51.709 & 47.887 & 26.970 & 4.976 & 0.640\\
600px & 53.423 & 49.675 & 25.669 & 4.290 & 0.498\\
800px & 54.744 & 50.560 & 24.033 & 3.046 & N/A\\
1000px & 55.163 & 49.466 & 20.061 & 2.071 & N/A\\\hline
\end{tabular}}
\caption{The impact of image size. Results are obtained with Mobile Parsing R-CNN (B s3x, test-tuned) on DensePose-COCO \textit{minival}. The N/A values correspond to tensor sizes that produced errors on mobile device}
\label{table:pixel}
\end{table*}

\begin{table*}[t!]
\centering
\resizebox{0.9\textwidth}{!}{
\begin{tabular}{ccccccccccc}
\hline
max \# of people & box AP & box APs & box APm & box APl & dp. AP & dp. APm & dp. APl & GPU FPS & CPU FPS & mobile CPU FPS\\\hline
1 & 83.110 & - & 83.389 & 83.173 & 54.329 & 48.203 & 54.765 & 27.508 & 5.859 & 0.684 \\
2 & 74.700 & 24.058 & 56.672 & 77.359 & 52.402 & 47.694 & 52.991 & 27.729 & 5.626 & 0.664 \\
3 & 71.508  & 16.357 & 54.621 & 76.280 & 52.324 & 47.973 & 52.905 & 26.767 & 5.584 & 0.638 \\
4 & 68.818 & 19.532 & 52.693 & 75.693 & 52.050 & 43.131 & 52.838 & 27.198 & 5.510 & 0.606 \\
5 & 66.756 & 20.252 & 53.543 & 74.807 & 51.468 & 44.154 & 52.501 & 27.732 & 5.443 & 0.603 \\\hline
\end{tabular}}
\caption{The impact of number of people in the frame on performance characteristics. Results are obtained with Mobile Parsing R-CNN (B s3x, test-tuned) on DensePose-COCO \textit{minival}. The shortest image side is 512 pixels}
\label{table:people}
\end{table*}

We evaluate the mobile model with Caffe2 runtime, running on a smartphone with ARM processor with 8 cores, 8 threads, and the highest core clock of 2600 MHz.

We use the deployment conversion tools provided by Detectron2 \cite{detectron2}. Specifically, the network is transferred first to ONNX format, then to Caffe2 format.

In general, two-stage models introduce numerous hyper-parameters. In case of test-time hyper-parameters, we found empirically, among many different options, that choosing $100$ instead of $1000$ region proposals per \textit{neck} level after non-maximum suppression (NMS) in RPN and changing IoU threshold in NMS from $0.5$ to $0.3$ leads to a significant boost of the model. Therefore later, we use this setup of the model and call it Mobile Parsing R-CNN (B s3x test-tuned).

We check the impact of the image size on the model (see Table~\ref{table:pixel}). The lower resolution of the image, the faster inference we get, but the reduction of image size results in a reduction of densepose AP. In the case of mobile inference, we apply the model on images with the shortest side of size 512 pixels, because it is the lowest resolution processed by the model during the training phase.

We are mostly interested in practical applications on the end-device with data fed straight from the device's camera. In this case, usually, the limited number of people appears in the frame. We test the model performance on filtered versions of COCO-DensePose minival partition, where the filtering is based on the maximum number of people in the image. The results can be seen in Table~\ref{table:people}. One can see that the fewer people are in the image, the better performance of the model in AP and FPS. Usually, the fewer people in the image, the more area each person occupies in the frame, which leads to more accurate predictions.

\subsection{Model quantization results}

Here we report the performance statistics of the model obtained using the quantization approach described in Section~\ref{quant}. Thanks to the quantization, we increased the speed of inference by a factor of two and decreased the model size by a factor of three. See exact values in Table~\ref{table:quant}.

\begin{table}[!t]
\centering
\resizebox{0.4\textwidth}{!}{
\begin{tabular}{cccccc}
\hline
weights type & model size & dp. AP & CPU FPS \\\hline
float32 & 13.8mb & 47.887 & 4.976 & \\
int8 & 4.3mb & 44.033 & 8.310 & \\\hline
\end{tabular}}
\caption{The effect of quantization. Results are obtained with Mobile Parsing R-CNN (B s3x, test-tuned) on DensePose-COCO minival. The shortest image side is 512 pixels}
\label{table:quant}
\end{table}
\section{Conclusion}

In this work we showed that it is possible to significantly compress and speed up models ($17\times$ model size reduction and $2\times$ latency improvement) for DensePose estimation task utilizing existing state-of-the-art solutions of this task's subproblems, achieving a good balance between speed, model size and average precision of the model. In the process, we performed an ablation study of 23 different backbones and detection pipeline characteristics, particularly applied for the DensePose task.
By optimizing different parts of R-CNN-like models, we achieved significant performance improvement compared to the baseline model.
We performed deployment of the final model to the mobile device, measured its performance, and discovered factors affecting it. 
The proposed architecture Mobile Parsing R-CNN is both fast and light-weight. Notably, the final model weighs 13.8MB and runs near real-time $\sim27$ FPS on Nvidia Tesla 1080Ti GPU, and $\sim1$ FPS on a mobile device using the only CPU. Using a runtime environment that utilizes mobile GPU or Neural Network acceleration hardware (NPUs), it would be trivial to get near-real-time performance on a mobile phone.
\newline\newline
\noindent\textbf{Acknowledgment.} The authors acknowledge the usage of the Skoltech CDISE HPC cluster Zhores for obtaining the results presented in this paper. This work was supported partially by the Ministry of Education and Science of the Russian Federation (Grant no. 14.756.31.0001).


{\small
\bibliographystyle{ieee_fullname}
\bibliography{egbib}
}

\end{document}